\documentclass[letterpaper, 10 pt, journal, twoside]{ieeeconf}   

\IEEEoverridecommandlockouts                              

\usepackage{amsmath}
\usepackage{yfonts}
\usepackage{booktabs}
\usepackage{amssymb}
\usepackage{graphicx}
\usepackage{multirow}
\usepackage{color}
\usepackage{float}
\usepackage{units}
\usepackage{footnote}
\usepackage{bm}
\makesavenoteenv{tabular}
\usepackage{pbox}
\makesavenoteenv{table}
\usepackage[all=normal,paragraphs=tight,floats=normal,mathspacing=normal,wordspacing=normal,charwidths=normal,mathdisplays=normal,leading=normal]{savetrees}
\newcommand{\camposetime}[1]{^{c_{#1}}\mathbf{T}_w}
\newcommand{\campose}[0]{^c\mathbf{T}_w}
\newcommand{\objposetime}[1]{^w\mathbf{T}_{o_{#1}}}

\newcommand{\pointidx}[2]{^{#1}\mathbf{X}_{#2}}
\newcommand{\point}[1]{{^{#1}\mathbf{X}}}
\newcommand{\projpoint}[1]{{^{#1}\mathbf{\Tilde{X}}}}

\newcommand{\pointh}[1]{^{#1}\mathbf{\bar{X}}}

\newcommand{\imgpointij}[2]{^{#1}\mathbf{x}_{#2}}
\newcommand{\twistcoord}[2]{^{#2}\bm{\xi}_{#1}}

\newcommand{\tildetwistcoord}[2]{^{#2}\bm{\Tilde{\xi}}_{#1}}

\newcommand{\pose}[2]{^{#1}\mathbf{T}_{#2}}
\newcommand{\posedet}[2]{^{#1}\mathbf{T}_{#2}^{d}}
\newcommand{\poseicp}[2]{^{#1}\mathbf{T}_{#2}^{r}}
\newcommand{\twistdet}[2]{^{#2}\bm{\xi}_{#1}^{d}}
\newcommand{\twisticp}[2]{^{#2}\bm{\xi}_{#1}^{r}}

\newcommand{\projmat}[0]{\bm{\Pi}}
\newcommand{\proj}[0]{p}

\newcommand{\rev}[0]{}

\title{
TwistSLAM++: Fusing multiple modalities for accurate \\dynamic semantic SLAM 
}

\author{Mathieu Gonzalez$^{1}$, Eric Marchand$^{2}$, Amine Kacete$^{1}$ and J\'erome Royan$^{1}$
\thanks{$^{1}$ Mathieu Gonzalez, Amine Kacete and Jerome Royan are with the Institute of Research and Technology b$<>$com, Rennes, France,
       {\tt\small \{mathieu.gonzalez,amine.kacete, jerome.royan\}@b-com.com}}%
\thanks{$^{2}$ Eric Marchand is with Univ Rennes, Inria, IRISA, CNRS, Rennes, France,
        {\tt\small Eric.Marchand@irisa.fr}}%
}

\begin{document}
\maketitle

\begin{abstract}
Most classical SLAM systems rely on the static scene assumption, which limits their applicability in real world scenarios. Recent SLAM frameworks have been proposed to simultaneously track the camera and moving objects. However they are often unable to estimate the canonical pose of the objects and exhibit a low object tracking accuracy. To solve this problem we propose TwistSLAM++, a semantic, dynamic, SLAM system that fuses stereo images and LiDAR information. 
Using semantic information, we track potentially moving objects and associate them to 3D object detections in LiDAR scans to obtain their pose and size. Then, we perform registration on consecutive object scans to refine object pose estimation. Finally, object scans are used to estimate the shape of the object and constrain map points to lie on the estimated surface within the bundle adjustment. We show on classical benchmarks that this fusion approach based on multimodal information improves the accuracy of object tracking.

\end{abstract}
~\\
\section{INTRODUCTION}

\label{sec:intro}
The goal of visual Simultaneous Localization and Mapping (SLAM) is to estimate the pose of a camera moving in space while simultaneously building a map of the environment. Classical approaches \cite{mur2017orb} assume the scene to be static, a condition that is rarely met in real world scenarios. To solve this problem some systems propose to mask out dynamic objects in images \cite{bescos2018dynaslam}. While this method enables SLAM in dynamic scenarios, it also loses an important piece of information for some applications. Indeed autonomous vehicles or augmented reality may need an estimate of the trajectory and pose of objects in the scene. Moreover some approaches mask out a priori dynamic objects that are in reality static (e.g. parked cars) which can hurt camera pose estimation accuracy.
\begin{figure}[ht]
    \centering
    \includegraphics[width=\columnwidth]{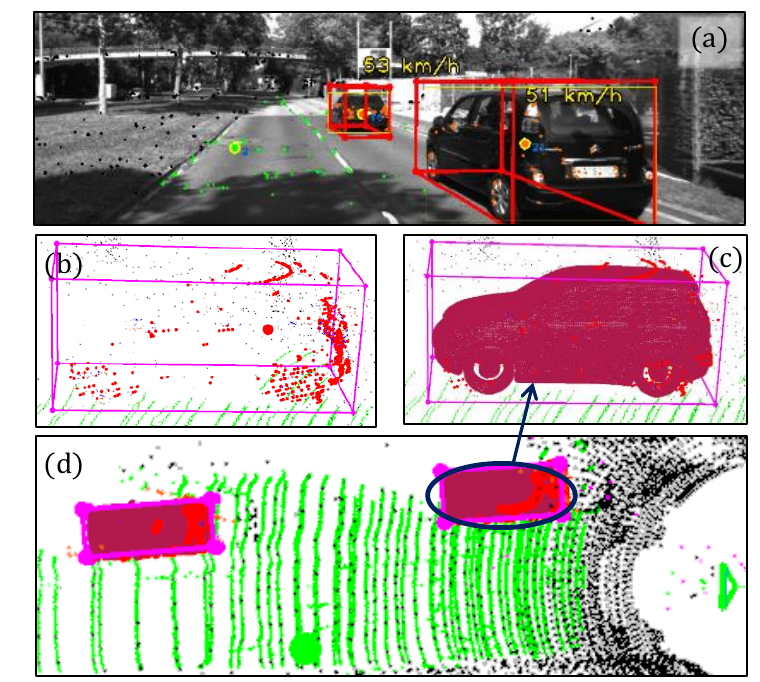}
    \vspace{-7mm}
    \caption[Example of tracked cars by TwistSLAM]{In our SLAM system we track object (here cars) moving in the scene. We show here (a) a frame with tracked cars, their speed and reprojected bounding boxes. (b) the bounding box and clustered LiDAR points (red) for the closest car. (c) the reconstruction of the car, using the approach of DSP-SLAM \cite{wang2021dsp}. (d) the map seen from above, with LiDAR points (black), road LiDAR points (green), tracked cars and the camera frustum (right).}

\vspace{-5mm}
    
    \label{fig:twistslam_first_fig}
\end{figure}
To solve this problem, systems such as \cite{bescos2021dynaslam, gonzalez2022twistslam}, have been designed to track both the camera and all moving objects. They show that they can accurately estimate camera poses in dynamic scenarios while estimating the trajectory and velocities of moving objects. However those approaches are often based solely on RGB information and are less precise than camera pose estimation. 
Furthermore, those approaches suffer from tracking drift that can not be corrected by loop closure.
Finally they do not have access to the canonical pose of the object but rather to its relative pose with respect to its initial pose. This can be limitating as relative pose alone is insufficient for some use cases, such as augmented reality applications that need an estimate of the object pose to seamlessly overlay virtual content on it. 
To solve those problems we propose to update our previous work TwistSLAM \cite{gonzalez2022twistslam} by integrating a 3D object detector based on LiDAR information. This detector is used to predict the pose and size of 3D bounding boxes corresponding to potentially moving objects in the scene. Associating detections to tracked objects allows us to have access to their canonical pose, i.e. their pose with respect to an a priori known object coordinate frame. Furthermore, we use consecutive poses to constrain the displacement of objects, thus reducing the drift. The obtained bounding boxes are then used to associate 3D LiDAR points to tracked clusters which serves two purposes. First they allow us to improve object tracking by feeding successive scans to a generalized ICP algorithm. The computed pose is then used as a constraint in the bundle adjustment (BA). Second, inspired by the work of DSP-SLAM \cite{wang2021dsp}, we use scans to fit a deep-learnt signed distance function (SDF) \cite{park2019deepsdf} that represents the object geometry. However contrary to DSP-SLAM we do not use the SDF to estimate the object pose as we already have a good estimate of it, but we rather use it to constrain the 3D map points of clusters to lie on the estimated mesh, similarly to our previous work \cite{gonzalez2021s3lam} which was restricted to planes. 
~\\
To summarize, our contributions are:
\begin{itemize}

\item A semantic SLAM system that can robustly estimate the pose of a camera in dynamic scenes.
\item A SLAM framework that can track multiple moving objects and estimate their canonical pose.
\item A SLAM system able to fuse 3D object pose estimation, object tracking and 3D registration results from LiDAR scans to reduce tracking drift.
\item A SLAM framework that uses the 3D reconstruction of object from LiDAR data to constrain the geometry of map points.
   
\end{itemize}

We evaluate our approach on sequences from the KITTI tracking datasets. We compare our results with state of the art dynamic SLAM systems DynaSLAM 2 \cite{bescos2021dynaslam} and TwistSLAM \cite{gonzalez2022twistslam} and show that we improve object tracking robustness and accuracy within our SLAM system.

The rest of the paper is described as follows. First we describe related work on  dynamic SLAM, object based SLAM and LiDAR based SLAM. Then, we rapidly recall the work of TwistSLAM and present the novelties of our approach: the use of a LiDAR based object detector, followed by scan registration and SDF fitting. Finally we demonstrate the benefits of our approach on multiple sequences from a public dataset.
\section{Related work: Dynamic and object based SLAM}
In this section we first present some semantic dynamic SLAM systems that tackle the problem of dynamic objects by tracking them \cite{runz2018maskfusion, bescos2021dynaslam, huang2020clustervo, zhang2020vdo, yang2019cubeslam} or use them as high level landmarks \cite{salas2013slam++, yang2019cubeslam}. For a additional resources on classical and semantic SLAM we refer the reader to \cite{cadena2016past, taketomi2017visual,gonzalez2022twistslam}. We also present SLAM systems that make use of LiDAR information and fuse multiple modalities to improve the accuracy and robustness of camera tracking.

DynaSLAM II \cite{bescos2021dynaslam} uses semantic information to detect objects. Object 3D points are represented in the object reference frame and used to estimate the object pose at all time by minimizing their reprojection error. 
TwistSLAM, \cite{gonzalez2022twistslam} creates a map of clusters corresponding to objects in the scene. Static objects are used for camera tracking while potentially dynamic objects are tracked. Furthermore, by using mechanical links between clusters, TwistSLAM constrains the velocity of objects to be coherent with the structure of the scene, improving object pose estimation.

Some approaches propose to detect objects in the scene to use them as high level landmarks. The first object based SLAM systems such as \cite{salas2013slam++, civera2011towards} require a specific object pose estimation algorithm \cite{rad2017bb8, gonzalez2021l6dnet} which limits their applicability in real world scenarios. More recent ones, however, are based on generic object detectors. 
Those approaches use quadrics \cite{nicholson2018quadricslam} or 3D bounding boxes \cite{yang2019cubeslam} to represent objects.
Some even more recent approaches have represented the geometry of objects more accurately using learning based approaches. NodeSLAM \cite{sucar2020nodeslam} optimizes detected object poses and shape, represented by an autoencoder. Object poses are then used in the SLAM system to estimate the camera pose.
DSP-SLAM \cite{wang2021dsp} optimizes the latent code of a deep learning based SDF \cite{park2019deepsdf} and uses it to estimate the pose of the object and to reconstruct the object shape. Object poses are then used to constrain camera pose estimation in the BA.

Finally, some approaches \cite{behley2018efficient, park2018elastic} have been using LiDAR scans instead of images as an input for SLAM: 
\cite{behley2018efficient} is a full SLAM system based only on LiDAR data, which represents the map using a set of surfels. 3D LiDAR points are transformed to the image plane using a spherical projection, yielding a so-called \textit{vertex map}. This map is used, together with a \textit{normal map} to estimate the updated current pose using point to plane registration after finding associations in the image plane. The current scan is then fused with the map to update it. Finally loop closure is performed. Virtual views are generated with the surfel map to compute the alignment with the current scan. After a verification step, a pose graph optimization is performed and used to update the surfel map.
Some approaches \cite{chen2019suma++, sun2018recurrent, chen2021moving, jeong2018multimodal} have also injected semantic information into LiDAR based SLAM systems, to improve pose estimation, for example by masking out moving objects.
SuMa++ \cite{chen2019suma++} improves on \cite{behley2018efficient} by integrating a CNN to segment LiDAR scans \cite{milioto2019rangenet++}. This allows them to obtain a higher level map. Furthermore semantic information is used to detect and remove surfels belonging to dynamic objects. It is also used to guide the ICP by weighting associated points.

\section{TwistSLAM++: Multimodal object tracking}
\begin{figure*}
    \centering
    \includegraphics[width = \textwidth]{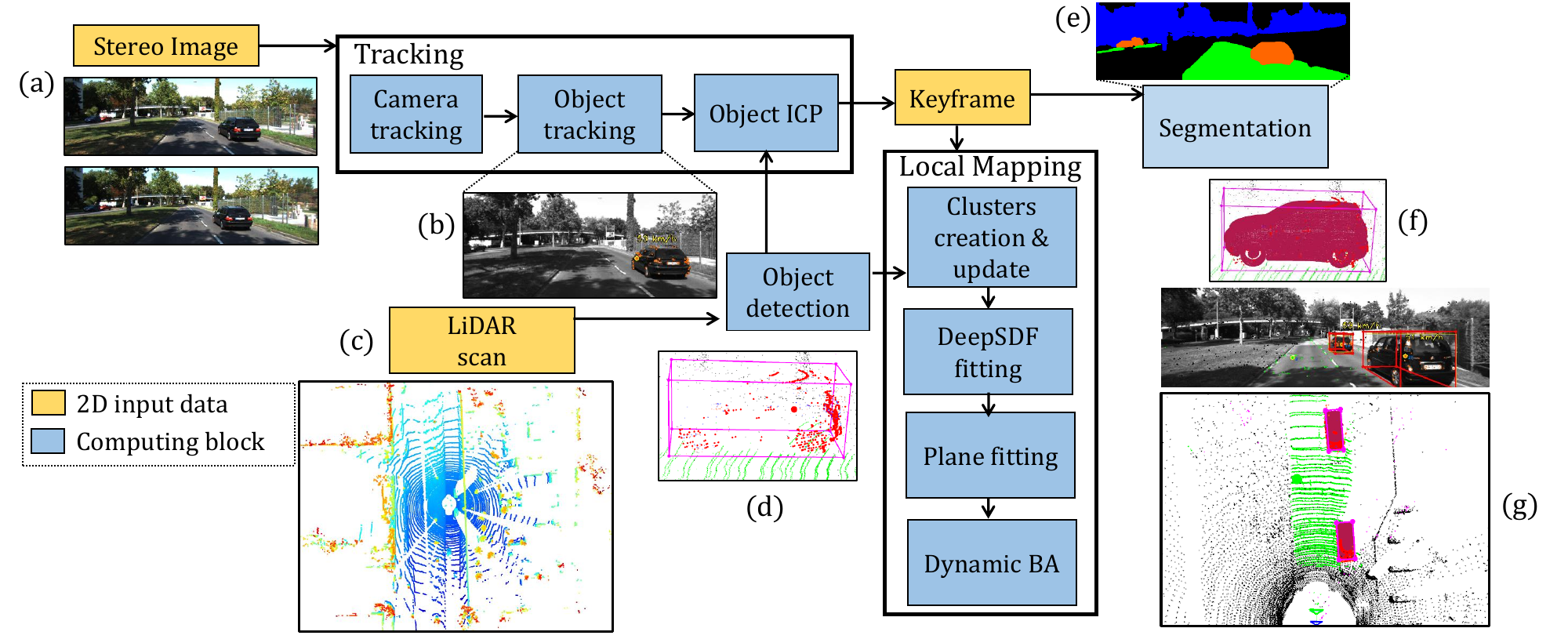}
    \vspace{-7mm}
    \caption{The pipeline of our approach: (a) keypoints are extracted from stereo images and used for camera tracking and (b) object tracking. (c) LiDAR scans are fed to a 3D object detector, allowing us to obtain (d) the 3D bounding box of objects in the scene and to cluster LiDAR points, which are then used in an ICP algorithm. (e) selected keyframes are segmented to create clusters that are augmented with clustered LiDAR points. (f) Object LiDAR points are then used to fit a per object sdf that constrains the geometry of clusters. (g) The trajectory and geometry of clusters is refined in the BA.}
    \vspace{-5mm}
    \label{fig:main_fig}
\end{figure*}

In this section we present our approach for which we show a pipeline in figure \ref{fig:main_fig}. 
Following the idea of the algorithms TwistSLAM \cite{gonzalez2022twistslam} and S$^3$LAM \cite{gonzalez2021s3lam} we use a panoptic neural network \cite{wu2019detectron2} to create a map of clusters corresponding to objects in the scene. Using points extracted from a priori static clusters (e.g. road, house, vegetation) we track the camera. Then, we use the points from remaining potentially dynamic clusters to track the objects. As we estimate the geometry of some objects (e.g. a plane for the road) we are able to constrain the velocity of tracked clusters with mechanical links. To improve this approach we chose to use LiDAR scans in several ways. First we feed them to a 3D object detection network that estimates the pose and size of objects in the scene. Second we use successive LiDAR scans corresponding to objects and register them to compute their relative pose. We inject both detected and registered poses as constrains in the BA, the first one being free from any drift and the second one more accurate. Third, we follow the work of \cite{wang2021dsp} and use DeepSDF \cite{park2019deepsdf} to fit a SDF to objects using LiDAR points. The SDF is then used in the BA to constrain the SLAM map points to lie on the object surface, thus improving the estimated map.

\subsection{Clusters creation}

To obtain a complete semantic map in which objects are uniquely identified we estimate the panoptic segmentation of images, using \cite{wu2019detectron2}. 
Similarly to \cite{gonzalez2021s3lam} we fuse 2D observations of a single 3D point to obtain its class and id. Doing so we obtain a semantic map and create a set of $K$ clusters $\mathcal{O} = \{O_{k}, k \in [1,K]\}$. A cluster is a set of 3D points corresponding to a single object, grouped according to their class and instance id. We split the set of clusters into two parts: static clusters $\mathcal{S}$ (such as \textit{road, building, ...}) and a priori dynamic clusters $\mathcal{D}$ (such as \textit{car, bike, human, bus, ...}).
A static cluster contains 3D points $\{\point{w}\}$ expressed in the world frame. On the other hand each dynamic cluster contains a set of 3D points $\{\point{o}\}$ expressed in the object coordinate frame, a set of poses $\{\pose{w}{o}\}$ and a set of twists expressed in the world coordinate frame $\{\twistcoord{o}{w}\}$ representing the cluster trajectory and velocity through time. 
The pose transforms points from the object coordinate frame to the world coordinate frame:
\begin{equation}
    \pointh{w} = \pose{w}{o} \,\pointh{o}
\end{equation}
where $\pointh{}$ denotes the homogeneous coordinates. The twist, of object $o$ at the $i^{th}$ timestamp, express in world coordinates transforms a pose between two timestamps:
\begin{equation}
    \pose{w}{o_{i+1}} = \exp(\twistcoord{o_i}{w}) \pose{w}{o_{i}} \label{eq:update_pose}
\end{equation}
where $\exp(.)$ is the exponential map \cite{blanco2010tutorial} of $se(3)$.
For simplicity in the remainder of this paper we will omit the object index $k$. For some clusters corresponding to a priori chosen classes (such as the road or the facade of a building) we estimate a 3D plane, represented by $\pi = (a,b,c,d)^\top$ with $||\pi||^2=1$, using its 3D points $\{\point{w}\}$. The plane follows the following equation: $\pi^{\top} \point{w} = 0$ and we estimate it using a SVD in a RANSAC loop.

\subsection{Dynamic SLAM}
Using static clusters we can robustly track the camera:
\begin{equation}
    E(\camposetime{i}) = \sum_{j \in \mathcal{S}}{\rho(||\imgpointij{i}{j}-\proj(\camposetime{i}, \pointidx{w}{j})||_{\Sigma_{i,j}^{-1}})}
    \label{eq:track_stat}
\end{equation}
where $\imgpointij{i}{j}$ is the 2D keypoint corresponding to the observation of $\pointidx{w}{j}$ in the $i^{th}$ frame, $\proj$ is the pinhole camera projection function, $\rho$ is a robust cost function (in our case Huber) \cite{malis2006experiments} and $\Sigma_{i,j}$ is the covariance matrix of the reprojection error. The camera pose is correctly estimated, even in dynamic scenes as it is only based on static parts of the scene. 

\subsection{Dynamic object tracking}
To track objects we match keypoints extracted from the current and the previous frame. This can be a challenging task as we do not initially know the movement of objects. To facilitate matching of keypoints and robustify it we estimate the optical flow between consecutive frames with a CNN \cite{teed2020raft}. Keypoints are searched in areas defined by their previous position updated with the optical flow. 
Furthermore we ensure that enough keypoints are available for tracking by extracting more keypoints from areas defined by objects bounding boxes. 
The keypoints are then used either to create new 3D points with stereo triangulation, which are added to existing clusters or used to create new clusters, or used to track the existing cluster.
The assumption in TwistSLAM \cite{gonzalez2022twistslam} is that many moving clusters can be represented as being linked to a static parent cluster with a specific mechanical link. To track the object we thus optimize the twist $\twistcoord{o_i}{w}$: 

\begin{equation}
\medmuskip=-2mu
\thinmuskip=-2mu
\thickmuskip=-1mu
    E(\twistcoord{o_i}{w}) =  \sum_j{\rho(||\imgpointij{i}{j}-\proj(\camposetime{i} \exp{(\projmat \twistcoord{o_i}{w})\objposetime{i-1}, \pointidx{o}{j}})||_{\Sigma_{i,j}^{-1}})}
    \label{eq:dyna_tracking}
\end{equation}
where $\bm{\Pi}$ is the projection operator that removes twists degrees of freedom, which constrains the twist according to the mechanical link, $\rho$ is the Huber robust estimator \cite{malis2006experiments} and $\Sigma_{i,j}$ is the covariance matrix of the reprojection error, which we estimate using the median absolute deviation (MAD) \cite{malis2006experiments}. For further information about the development of the projection operator we refer the reader to \cite{gonzalez2022twistslam}. We then update the object pose using equation (\ref{eq:update_pose}).

We optimize this cost function with the Levenberg-Marquardt algorithm on matches found between consecutive frames. We then use the estimated pose to project 3D map points into the current frame, find new matches and optimize again the cost function.
\subsection{Injecting LiDAR scans in TwistSLAM}
To improve the accuracy of object tracking we propose to use LiDAR scans, taken at each timestamp and  processed in multiple ways. LiDAR scans are loaded by the tracking thread and tranformed from the LiDAR to the camera coordinate frame. 

The first way we process the scans is by using a 3D object detection network (namely 3DSSD \cite{yang20203dssd}). For each scan this network yields a set of detected objects, with their corresponding size and pose, denoted $\posedet{c_i}{o_{i}}$ where $d$ stands for detection for the estimation at the $i^{th}$ frame.
We associate clusters created in our SLAM system to object detections by minimizing the 3D distance between the box center and the cluster centroid. A detection is valid if its distance to the cluster centroid is lower than 2 meters. 
We then use consecutive detections to compute the relative twist $\twistdet{o_i}{w}$ linking two poses:
\begin{equation}
    \twistdet{o_i}{w} = \log(\posedet{c_{i+1}}{o_{i+1}} \, (\posedet{c_{i}}{o_{i}})^{-1})
\end{equation}
The estimation of this twist has the advantage of being free from any drift. It can thus be used to limit the drift accumulated during tracking, similarly to the action of a loop closing step for camera tracking. The drawback however is that it is more noisy than keypoints based tracking. Indeed, the detector was trained to detect 3D objects rather to accurately estimate their pose

We show in section \ref{section:ba} how to inject this estimated twist in the BA to improve object tracking.

One of the main limitations of TwistSLAM \cite{gonzalez2022twistslam} is the lack of canonical pose for objects. Indeed, when an object is first created, its pose is initialized with an identity matrix for the rotation, and the centroid of the cluster for the translation. Thus this pose does not relate to the pose of objects in their canonical coordinate frame. Using 3D object detection we can estimate the initial object pose. This initial pose is then updated by object tracking and by the BA, using estimated twists.
We also fuse the estimated dimensions of the detection using the median of each dimension for robustness. 
Both the pose and the dimensions allow us to estimate a 3D bounding box for clusters.

\par
Using this box we can associate 3D LiDAR points to clusters.
We thus obtain at each timestamp a precise 3D scan of each object in the scene.
We apply a generalized ICP algorithm \cite{segal2009generalized} to compute the transformation between consecutive timestamps, that we denote $\poseicp{o_{i+1}}{o_i}$ where $r$ stands for registration.
This transformation can be decomposed as:
\begin{equation}
    \poseicp{o_{i+1}}{o_{i}} = \poseicp{o_{i+1}}{w} \, (\poseicp{o_{i}}{w})^{-1}
\end{equation}
which allows us to compute the corresponding twist in the world coordinate frame:
\begin{equation}
    \twisticp{o_{i}}{w} = \log(\poseicp{o_{i+1}}{o_{i}})
\end{equation}
This twist has the advantage of being accurate compared to keypoint based twists, particularly when keypoints are difficult to extract (e.g. on small, far or textureless objects). We show in section \ref{section:ba} how to inject it in the global BA.
\par
Finally, to improve the estimation of plane parameters we propose to use LiDAR points that are more precise, denser and cover more space than triangulated points. We transform LiDAR scans to world coordinates using the estimated camera camera pose and project them into the segmented image. If their class is a priori planar and their score higher than a threshold, we append them to the cluster and use them for plane fitting. We apply this strategy to the class \textit{road}.

\subsection{Estimating clusters geometry}
The geometry of objects is an important property that we can inject in a SLAM system to improve the accuracy of 3D mapping. To estimate it we use clustered LiDAR points, that are precise and apply the method developed in DSP-SLAM \cite{wang2021dsp}. 
DSP-SLAM uses DeepSDF \cite{park2019deepsdf} to represent the geometry of an object with a signed distance function generated from its latent code vector:
\begin{equation}
    G(\point{o}, \mathbf{z}) = s
\end{equation}
where $s$ is the SDF value computed at the 3D points position $\point{o}$ and $\mathbf{z} \in \mathbb{R}^{64}$ is the latent code representing the object shape.
They optimize the latent code, object pose and scale so that the generated geometry tightly fits the object scan.
Doing so it is possible to reconstruct a realistic watertight mesh and use the object poses as constrains in the BA.
As we already have a good estimate of the object canonical poses we propose to apply their algorithm on LiDAR points not to refine the pose but rather to refine the SLAM 3D points.
We use clustered LiDAR scans to fit the latent code $\mathbf{z}$ similarly to DSP-SLAM. However contrary to DSP-SLAM we keep the object pose and scale fixed, their values being set using our own estimate of the object pose and length. 
Then we seek to constrain 3D map points so that they lie on the object estimated surface. As we have an estimate of the SDF value and of its derivative with respect to the 3D points position we can apply a gradient descent algorithm to project points on the surface. At the $k^{th}$ step of the algorithm we have:
\begin{equation}
    \point{o}^{(k+1)} = \point{o}^{(k)} - \alpha^{(k)} \frac{\partial G(\point{o}^{(k)}, \mathbf{z})}{\partial \point{o}^{(k)}}
\end{equation}
where $\alpha^{(k)}$ is the step size, the point initial value is $\point{o}^{(0)} =\point{o}$ and the derivative of $G$ is obtained through back propagation. This process is repeated for 10 steps to obtain projected points that we denote $\projpoint{o}$. Projected points will then be used as anchors in the bundle adjustment to constrain 3D points to be coherent with the estimated geometry. LiDAR scans, which are usually more precise than points triangulated from stereo images are thus used to constrain the map.

\subsection{Dynamic Bundle Adjustment}
\label{section:ba}
In TwistSLAM \cite{gonzalez2022twistslam} the bundle adjustment is used to refine all object and camera poses as well as all static and dynamic point positions. Furthermore it links consecutive poses so that their twists follow a constant velocity model. Doing so dynamic points are used to improve camera pose estimation. In our new BA we improve upon \cite{gonzalez2022twistslam} by adding new regularization terms, taking into account LiDAR scans processed in three ways: using a 3D object detection network, an iterative closest point algorithm and a deep signed distance function fitting.

Our bundle adjustment cost function can be written as follows:
\begin{equation}
\begin{split}
    E(\{\twistcoord{o}{w}, \campose, \point{w},  \point{o}\}) = \sum_{i,j} e_{stat}^{i,j}+\sum_{i,j}e_{dyna}^{i,j}+ \\ \sum_{i}e_{const}^{i}
    + \sum_{i}e_{reg}^{i} + \sum_{i}e_{det}^{i} + \sum_{j}e_{geo}^{j} \label{eq:ba}
\end{split}
\end{equation}
where $e_{stat}^{i,j}$ is the classical static reprojection error:
\begin{equation}
\nonumber
    e_{stat}^{i,j} = \rho(||\imgpointij{i}{j} - \proj(\camposetime{i}, \pointidx{w}{j})||_{\Sigma_{i,j}^{-1}})
\end{equation}
$e_{dyna}^{i,j}$ is a dynamic reprojection error: 
\begin{equation}
\nonumber
   e_{dyna}^{i,j} = \rho(||\imgpointij{i}{j} - \proj(\campose \exp(\projmat\twistcoord{o_i}{w})\objposetime{i}, \point{o})||_{\Sigma_{i,j}^{-1}})
\end{equation}
where $\Sigma_{i,j}^{-1}$ is estimated using the MAD as in equation (\ref{eq:dyna_tracking}).
$e_{const}^{i}$ is a constant velocity model that penalizes twists variations by linking 3 consecutive poses:
\begin{equation}
\nonumber
    e_{const}^{i} = \rho(||\projmat\tildetwistcoord{o_{i+1}}{w}-\projmat\tildetwistcoord{o_i}{w}||_{\mathbf{W}_{const}})
\end{equation}
where $\mathbf{W}_{const}$ is a diagonal weight matrix used to balance the errors, tuned experimentally,  $\tildetwistcoord{o_{i+1}}{w}$ is the twist linking the poses $\exp(\projmat\twistcoord{o_i}{w})\objposetime{i}$ and $\exp(\projmat\twistcoord{o_{i+1}}{w})\objposetime{i+1}$ and $\tildetwistcoord{o_{i}}{w}$ is the twist linking the poses $\exp(\projmat\twistcoord{o_{i-1}}{w})\objposetime{i-1}$ and $\exp(\projmat\twistcoord{o_{i}}{w})\objposetime{i}$. Those twists are computed using the logmap from $\mathrm{SE}(3)$ to $\mathrm{se}(3)$ defined in \cite{blanco2010tutorial} and can be written for $\tildetwistcoord{o_{i+1}}{w}$ as:
\begin{equation}
\nonumber
    \tildetwistcoord{o_{i+1}}{w} = \log(\exp((\projmat\twistcoord{o_{i+1}}{w})\objposetime{i+1})(\exp(\projmat\twistcoord{o_i}{w})\objposetime{i})^{-1})
\end{equation}
The error $e_{det}^{i}$ penalizes the difference with twists estimated from the object detection network:
\begin{equation}
\nonumber
    e_{det}^{i} = \rho(||\projmat\twistdet{o_{i+1}}{w}-\projmat\tildetwistcoord{o_{i+1}}{w}||_{\mathbf{W}_{det}})
\end{equation}
where $\mathbf{W}_{det}$ is a diagonal weight matrix.
Similarly, the error $e_{reg}^{i}$ penalizes the difference with twists estimated by registering consecutive point clouds:
\begin{equation}
\nonumber
    e_{reg}^{i} = \rho(||\projmat\twisticp{o_{i+1}}{w}-\projmat\tildetwistcoord{o_{i+1}}{w}||_{\mathbf{W}_{reg}})
\end{equation}
where $\mathbf{W}_{reg}$ is a diagonal weight matrix.
~\\
Finally, the residual $e_{geo}^{j}$ constrains points to lie on the estimated geometry surface by penalizing the difference between the position of 3D points and of their projected counterpart:
\begin{equation}
    e_{geo}^{j} = \rho(||\projpoint{o}_j-\point{o}_j||_{\mathbf{W}_{geo}})
\end{equation}
where $\mathbf{W}_{geo}$ is a diagonal weight matrix. To avoid corrupting map points due to wrong pose estimations or wrong projections, we only apply this constraint if its value is below some threshold.
Note that we could also directly use the DeepSDF function $G(\point{o}, \mathbf{z})$ to compute the value of the residual and to obtain the jacobian via back propagation. 
~\\
This equation refines all camera and object poses as well as all 3D points.
To optimize it in real time we apply the Schur trick as the Hessian is sparse \cite{bescos2021dynaslam}. 
For the management of keyframes, we adopt the same strategy as TwistSLAM \cite{gonzalez2022twistslam}, with temporal and spatial keyframes. 

\section{Experiments}
In this section we present the experiments we conducted to test our approach. We evaluate both the accuracy of the camera pose estimation and of the object pose estimation. 
\subsection{Experiments details}
\textbf{Datasets.}
We evaluate our approach on the KITTI \cite{geiger2012we} tracking dataset as it contains both camera and object trajectories groundtruth.
\rev{Points segmented by the network as the \textit{unknown} class are considered to be static, as the dynamic classes are often correctly segmented.}
~\\
\textbf{Metrics.}
To evaluate the accuracy of camera pose estimation we compute the translation and rotation parts of the Relative Pose Error (RPE) \cite{zhang2018tutorial}, similarly to previous works.  
We also evaluate the object pose accuracy using the Absolute Translation Error (ATE) and RPE. Furthermore we evaluate precision of objects 3D bounding boxes estimations by computing the MOTP, similarly to \cite{bescos2021dynaslam} using KITTI evaluation tools.
We evaluate the true positive rate (TP) and the MOTP using the projected 3D bounding box (2D), in bird view (BV) and in 3D. Those evaluations are done in the \textit{easy} setting as in \cite{bescos2021dynaslam}.
\subsection{Camera pose estimation}
In this subsection we evaluate the accuracy of our camera pose estimation which can be seen in table \ref{table:ego}. 
As we obtain almost exactly the same results as TwistSLAM we only show here the results on some sequences.
This is not surprising as the sequences only exhibit a mild amount of dynamicity, that is well dealt even by non dynamic approaches \cite{mur2017orb}. Furthermore in this approach we rather focused on the accuracy of object tracking, that we improve compared to state of the art, as shown in the following paragraph.

\begin{table*}
\caption{Camera pose estimation comparison on the Kitti tracking dataset.}
\vspace{-2mm}
\label{table:ego}
\centering
\resizebox{\textwidth}{!}{%
\begin{tabular}{@{}c|cc|cc|ll|cc|cc@{}}
\toprule
\multirow{2}{*}{seq} & \multicolumn{2}{c|}{ORB-SLAM2 \cite{mur2017orb}} & \multicolumn{2}{c|}{DynaSLAM \cite{bescos2018dynaslam}} & \multicolumn{2}{c|}{DynaSLAM2 \cite{bescos2021dynaslam}} & \multicolumn{2}{c|}{TwistSLAM \cite{gonzalez2022twistslam}}  & \multicolumn{2}{c}{Ours} \\ \cmidrule(l){2-11} 
 & RPE$_t$ (m/f) & RPE$_R$ (°/f) & RPE$_t$ (m/f) & RPE$_R$ (°/f) & RPE$_t$ (m/f) & RPE$_R$ (°/f) & RPE$_t$ (m/f) & RPE$_R$ (°/f) & RPE$_t$ (m/f) & RPE$_R$ (°/f) \\ \midrule
00 & \bf{0.04} & 0.06 & \bf{0.04} & 0.06 &\bf{0.04} & 0.06 & \bf{0.04} & \bf{0.05} & \bf{0.04} & \bf{0.05} \\
02 & 0.04 & 0.03 & 0.04 & 0.03 & 0.04 & \bf{0.02} & 0.03 & 0.03& 0.03 & \bf{0.02} \\
03 & 0.07 & 0.04 & 0.07 & 0.04 & 0.06 & 0.04 & 0.06 & \bf{0.02} & 0.06 & 0.03\\
04 & 0.07 & 0.06 & 0.07 & 0.06 & 0.07 & 0.06 & 0.06 & \bf{0.04}& 0.06 & \bf{0.04} \\
05 & 0.06 & 0.03 & 0.06 & 0.03 & 0.06 & 0.03 & 0.06 & \bf{0.02} & 0.06 & \bf{0.02} \\
10 & \bf{0.07} & 0.04 & \bf{0.07} & 0.04 & \bf{0.07} & 0.03 & \bf{0.07} & 0.03& \bf{0.07} & \bf{0.02} \\
11 & 0.04 & 0.03 & 0.04 & 0.03 & 0.04 & 0.03 & \bf{0.03} & \bf{0.02}& \bf{0.03} & \bf{0.02} \\
14 & \bf{0.03} & 0.08 & \bf{0.03} & 0.08 & \bf{0.03} & 0.08 & \bf{0.03} & \bf{0.06}& \bf{0.03} & \bf{0.06} \\
18 & 0.05 & 0.03 & 0.05 & 0.03 & 0.05 & \bf{0.02} & 0.04 & \bf{0.02}& 0.04 & \bf{0.02} \\\midrule
mean & 0.052 & 0.044 & 0.052 & 0.046 & 0.051 & 0.041 & \bf{0.041} & \bf{0.032} & \bf{0.041} & \bf{0.032} \\ 
\bottomrule
\end{tabular}%
}
\end{table*}
\subsection{Object pose estimation.}
In this paragraph we evaluate the accuracy of our object pose estimation, we show the results in tables \ref{table:obj_pose} and \ref{table:obj_pose_motp}. 
As we can see in table \ref{table:obj_pose} we obtain better results in terms of object tracking accuracy for the ATE and RPE compared to \cite{bescos2021dynaslam} and \cite{gonzalez2022twistslam}. We particularly improve the RPE for both the rotation and translation, which shows that adding constrains from processed LiDAR data reduces tracking drift. The TP and MOTP metrics in table \ref{table:obj_pose_motp} are improved compared to state of the art on average. On most sequences they show very similar scores to TwistSLAM, which can be expected as the MOTP metrics only require an overlap of 0.25 for a detection to be positive, thus an improvement of even tens of centimeters on the pose may not translate to new positive detections. A way to do so would be to decrease the number of points required for tracking to track objects for longer periods. This however is out of scope of our work as we focus on improving tracking accuracy. On some sequences however we obtain lower 3D and birdview MOTP for a similar 2D MOTP. Those differences are mainly due to wrong pose estimates that happen when the cluster is first created far from the camera. Those differences can also be explained by the fact that TwistSLAM uses the grountruth initial bounding box of objects, which is noise-free while we use a 3D object detector. Finally on some sequences (such as 11/35 and 20/0) the additional LiDAR information allows us to improve tracking stability and thus to track on longer trajectories, increasing the MOTP. 

\begin{figure*}
    \centering
    \includegraphics[width=1.0\textwidth]{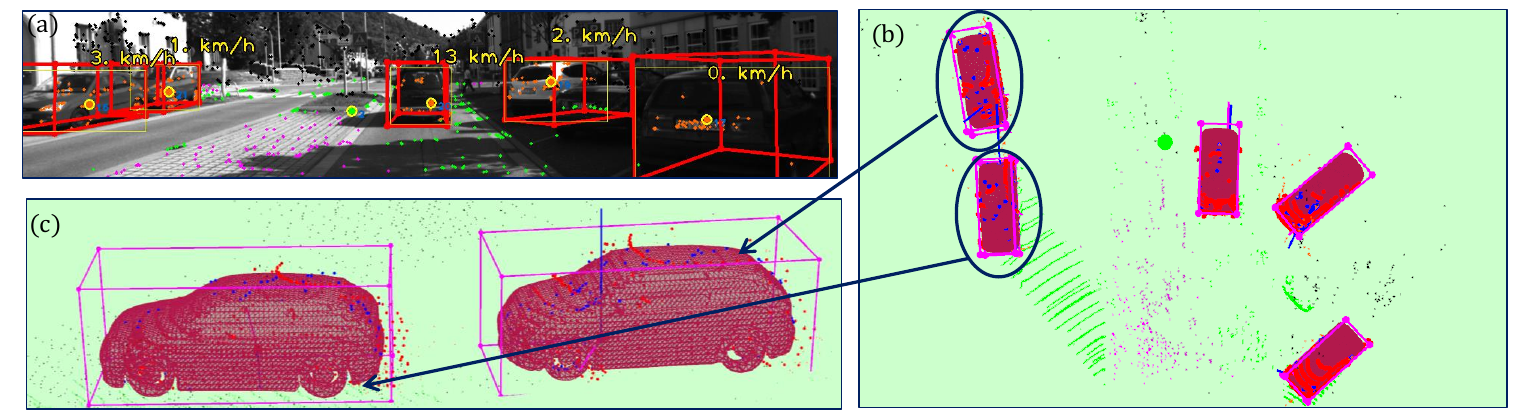}
    \caption[Example of tracked cars by TwistSLAM]{(a) Frame with detected objects, bounding boxes and speed.(b) Map with tracked objects, seen from above. (c) Mesh of reconstructed cars with bounding boxes, LiDAR points (red) and projected points on the mesh (blue).}

    \label{fig:quali}
\end{figure*}

\begin{table*}
\centering
\caption{Object pose estimation comparison on the Kitti tracking dataset. ATE is in m, RPE$_t$ in m/m, RPE$_R$ in °/m}
\label{table:obj_pose}
\begin{tabular}{@{}cccccccccc@{}}
\toprule
 & \multicolumn{3}{|c||}{DynaSLAM 2 \cite{bescos2021dynaslam}} & \multicolumn{3}{c||}{TwistSLAM} & \multicolumn{3}{c}{TwistSLAM++}\\ \midrule
\multicolumn{1}{c|}{seq / obj. id / class} & ATE & RPE$_t$ & \multicolumn{1}{c|}{RPE$_R$} &ATE & RPE$_t$ & \multicolumn{1}{c|}{RPE$_R$} &ATE & RPE$_t$ & \multicolumn{1}{c}{RPE$_R$} \\ \midrule
\multicolumn{1}{c|}{03 / 1 / car}&0.69&0.34&\multicolumn{1}{c|}{1.84}&0.31&\textbf{0.10}&\multicolumn{1}{c|}{0.28}& \textbf{0.23}&0.11&\multicolumn{1}{c}{\textbf{0.19}}\\
\multicolumn{1}{c|}{05 / 31 / car} & 0.51 & 0.26 &  \multicolumn{1}{c|}{13.5} & 0.58 & 0.35 & \multicolumn{1}{c|}{\textbf{0.19}} & \textbf{0.09} & \textbf{0.07} & \multicolumn{1}{c}{0.28}\\
\multicolumn{1}{c|}{10 / 0 / car} & 0.95 & 0.40 & \multicolumn{1}{c|}{2.84} & 0.77 & 0.21 & \multicolumn{1}{c|}{1.98} &  \textbf{0.05} & \textbf{0.10} & \multicolumn{1}{c}{\textbf{0.96}}\\
\multicolumn{1}{c|}{11 / 0 / car} & 1.05 & 0.43 & \multicolumn{1}{c|}{12.51} &0.17 & \bf{0.23} & \multicolumn{1}{c|}{0.23} & \bf{0.15} & 0.28 & \multicolumn{1}{c}{\bf{0.21}} \\
\multicolumn{1}{c|}{11 / 35 car} & 1.25 & 0.89 & \multicolumn{1}{c|}{16.64} & 0.10 & 0.03 & \multicolumn{1}{c|}{0.11} & 0.11 & \textbf{0.02} & \multicolumn{1}{c}{\textbf{0.09}} \\
\multicolumn{1}{c|}{18 / 2 / car} & 1.10 & 0.30 & \multicolumn{1}{c|}{9.27} & \textbf{0.21} & 0.27 & \multicolumn{1}{c|}{0.66} & \textbf{0.29} & \textbf{0.09} & \multicolumn{1}{c}{\textbf{0.32}} \\
\multicolumn{1}{c|}{18 / 3 / car} & 1.13 & 0.55 & \multicolumn{1}{c|}{20.05} &  0.15 & 0.21 & \multicolumn{1}{c|}{0.56} &  \bf{0.13} & \bf{0.10} & \multicolumn{1}{c}{\bf{0.37}} \\
\multicolumn{1}{c|}{19 / 63 / car} & 0.86 & 1.45 & \multicolumn{1}{c|}{48.80} &  \bf{0.28} & 2.17 & \multicolumn{1}{c|}{\bf{1.08}} & 0.34 & \bf{0.21} & \multicolumn{1}{c}{\bf{0.31}} \\
\multicolumn{1}{c|}{19 / 72 / car} & 0.99 & 1.12 & \multicolumn{1}{c|}{3.36}& 0.16 & 0.05 & \multicolumn{1}{c|}{\bf{0.34}} &  \bf{0.09} & \bf{0.03} & \multicolumn{1}{c}{0.37} \\
\multicolumn{1}{c|}{20 / 0 / car} & 0.56 & 0.45 & \multicolumn{1}{c|}{1.30} & \bf{0.17} & \bf{0.20} & \multicolumn{1}{c|}{0.72} &  0.30 & 0.21 & \multicolumn{1}{c}{\bf{0.35}} \\
\multicolumn{1}{c|}{20 / 12 / car} & 1.18 & 0.40 & \multicolumn{1}{c|}{6.19} &  \bf{0.24} & \bf{0.20} & \multicolumn{1}{c|}{1.54} &  0.80 & 0.54 & \multicolumn{1}{c}{\bf{0.64}}  \\
\multicolumn{1}{c|}{20 / 122 / car} & 0.87 & 0.72 & \multicolumn{1}{c|}{5.75} & 0.17 & \bf{0.02} & \multicolumn{1}{c|}{0.07} & \bf{0.16} & \bf{0.02} & \multicolumn{1}{c}{\bf{0.06}} \\ \midrule
\multicolumn{1}{c|}{mean} & 0.93& 0.61 & \multicolumn{1}{c|}{11.83} &{0.26} & {0.32} & \multicolumn{1}{c|}{{0.68}} & \bf{0.23} & \bf{0.15} & \multicolumn{1}{c}{\bf{0.35}} \\ \bottomrule
\end{tabular}%
\end{table*}
\begin{table*}
\caption{Object pose estimation comparison on the Kitti tracking dataset. TP and MOTP are in \%. }
\label{table:obj_pose_motp}
\resizebox{\textwidth}{!}{%
\begin{tabular}{@{}ccccccccccccccccccc@{}}
\toprule
 & \multicolumn{6}{c|}{DynaSLAM 2 \cite{bescos2021dynaslam}} & \multicolumn{6}{c|}{TwistSLAM} & \multicolumn{6}{c}{TwistSLAM++}\\ \midrule
\multicolumn{1}{c|}{seq / obj. id / class} & 2D TP & 2D MOTP & BV TP & BV MOTP & 3D TP & \multicolumn{1}{c|}{3D MOTP} &  2D TP & 2D MOTP & BV TP & BV MOTP & 3D TP & \multicolumn{1}{c|}{3D MOTP} & 2D TP & 2D MOTP & BV TP & BV MOTP & 3D TP & 3D MOTP \\ \midrule
\multicolumn{1}{c|}{03 / 1 / car}&50.0&\textbf{71.79}&39.34&56.61&38.53&\multicolumn{1}{c|}{48.20}& \textbf{58.02} & 60.00 & \textbf{58.02} & \textbf{60.00}  & \textbf{58.02} & \multicolumn{1}{c|}{\textbf{45.00}}  & \textbf{56.79} & 60.00 & \textbf{56.79} & \textbf{60.00}  & \textbf{56.79} & \textbf{60.00}\\
\multicolumn{1}{c|}{05 / 31 / car} & 28.96 & \textbf{60.30} & 14.48 & \textbf{46.84} & 11.45 & \multicolumn{1}{c|}{34.20} &  \textbf{30.84} & 35.00 &\textbf{30.84} & 35.00 & \textbf{30.84} & \multicolumn{1}{c|}{\textbf{35.00}} & 30.00  & 26.64 & 16.32 & 14.95 & 16.10 & 14.49\\
\multicolumn{1}{c|}{10 / 0 / car} & \bf{81.63} & \bf{73.51} & \bf{70.41} & \bf{47.60} & \bf{68.37} & \multicolumn{1}{c|}{\bf{40.28}} & 7.20 & 3.70 & 6.10 & 3.10 & 5.80 & \multicolumn{1}{c|}{2.80}& 6.48 & 10.00 & 6.48 & 10.00 & 6.48 & 10.00\\
\multicolumn{1}{c|}{11 / 0 / car} & \bf{72.65} & \bf{74.78} & \bf{61.66} & \bf{50.74} & \bf{52.28} & \multicolumn{1}{c|}{\bf{47.35}} &  29.61 & 32.50 & 29.61 & 32.50 & 29.61 & \multicolumn{1}{c|}{32.50}&  26.82 & 30.00 & 26.82 & 30.00 & 26.82 & 30.00 \\
\multicolumn{1}{c|}{11 / 35 car} &  53.17 & 65.25 & 19.05 & 31.95 & 6.35 & \multicolumn{1}{c|}{26.02} &  \textbf{65.00} & \textbf{67.50} & \textbf{65.00} & \textbf{67.50} & \textbf{65.00} & \multicolumn{1}{c|}{\textbf{67.50}}&  \textbf{73.75} & \textbf{77.50} & \textbf{73.75} & \textbf{77.50} & \textbf{73.75} & \textbf{77.50} \\
\multicolumn{1}{c|}{18 / 2 / car} &  \textbf{86.36} & 74.81 & 67.05 & 45.47 & 62.12 & \multicolumn{1}{c|}{34.80} & 84.67 & \textbf{87.50} & \textbf{84.67} & \textbf{87.50} & \textbf{84.67} & \multicolumn{1}{c|}{\textbf{87.50}} & 85.18 & \textbf{87.50} & \textbf{85.18} & \textbf{87.50} & \textbf{85.18} & \textbf{87.50}\\
\multicolumn{1}{c|}{18 / 3 / car} & \bf{53.33} & \bf{70.94} & 21.75 & \bf{41.45} & 16.84 & \multicolumn{1}{c|}{\bf{35.80}} &  28.19 & 30.00 & \bf{28.19} & 30.00 & \bf{28.19} & \multicolumn{1}{c|}{30.00}&  21.83 & 25.00 & {21.83} & 25.00 & {21.83} & 25.00 \\
\multicolumn{1}{c|}{19 / 63 / car} &  35.26 & 63.50 & 29.48 & 45.69 & 26.48 & \multicolumn{1}{c|}{33.89} & \bf{65.93} & \bf{70.00} & \bf{65.93} & \bf{70.00} & 36.26 & \multicolumn{1}{c|}{20.64} & \bf{65.93} & \bf{70.00} & \bf{65.93} & \bf{70.00} & \bf{65.93} & \bf{70.00} \\
\multicolumn{1}{c|}{19 / 72 / car} & \bf{29.11} & \bf{62.59} & \bf{29.43} & \bf{55.48} & \bf{29.43} & \multicolumn{1}{c|}{\bf{39.81}} & 16.92 & 20.00 & 16.92 & 20.00 & 16.92 & \multicolumn{1}{c|}{20.00}& 5.38 & 10.00 & 5.38 & 10.00 & 5.38 & 10.00 \\
\multicolumn{1}{c|}{20 / 0 / car} & 63.68 & 78.54 & 43.78 & 45.00 & 31.84 & \multicolumn{1}{c|}{46.15} &  {84.75} & {87.50} & {84.75} & {87.50} & {84.75} & \multicolumn{1}{c|}{{87.50}} &  \bf{93.22} & \bf{97.50} & \bf{93.22} & \bf{97.50} & \bf{93.22} & \bf{97.50}\\
\multicolumn{1}{c|}{20 / 12 / car} & \bf{42.77} & \bf{76.77} & \bf{37.64} & \bf{49.29} & \bf{36.23} & \multicolumn{1}{c|}{\bf{40.81}} & 14.24 & 17.5 & 13.91 & 17.45 & 13.04 & \multicolumn{1}{c|}{17.25}& 32.75 & 37.5 & 32.75 & 37.5 & 32.75 & 37.5 \\
\multicolumn{1}{c|}{20 / 122 / car} & 34.90 & 78.76 & 34.51 & 48.05 & 29.02 & \multicolumn{1}{c|}{44.43} & \bf{84.94} & \bf{87.50} & \bf{84.94} & \bf{87.50} & \bf{84.94} & \multicolumn{1}{c|}{\bf{87.50}} & \bf{84.94} & \bf{87.50} & \bf{84.94} & \bf{87.50} & \bf{84.94} & \bf{87.50} \\ \midrule
\multicolumn{1}{c|}{mean}  & \bf{55.15} & \bf{70.96} & 39.05& 47.01  & 34.08 & \multicolumn{1}{c|}{39.31 } & 45.53 & 49.89 & 47.41 & {49.84} & {44.84} & \multicolumn{1}{c|}{{43.18}} &  49.42 & 51.6 & \bf{47.45} & \bf{50.62} & \bf{47.43} & \bf{50.58} \\ \bottomrule
\end{tabular}%
}
\end{table*}
In addition to the first figure, we also show qualitative results in figure  \ref{fig:quali}.  

\section{Conclusion}
In this paper we proposed TwistSLAM++, an improvement over our previous work TwistSLAM, able to track the camera in dynamic scenes and estimate the canonical pose of all potentially moving objects. By injecting LiDAR data in our pipeline, we estimate the canonical pose and size of objects using a 3D object detection network. Then, we use consecutive clustered LiDAR scans to accurately compute their relative pose using an ICP algorithm, allowing us to further constrain their movement. Finally, we use object scans to estimate the 3D geometry of objects and use it to constrain the 3D position of map points. We show that adding those constraints from a new sensor allows us to improve object pose accuracy compared to the state of the art.




\bibliographystyle{IEEEtran} 
\bibliography{IEEEabrv,IEEEexample}
\end{document}